\documentclass[letterpaper]{article}
\usepackage{iccc}
\usepackage{times}
\usepackage{helvet}
\usepackage{courier}
\usepackage{xcolor}
\usepackage{url}
\usepackage{hyperref}
\usepackage{graphicx}
\usepackage{amsmath,amssymb}
\usepackage{color}
\usepackage{epsfig}
\usepackage{pdfpages}
\usepackage{caption}
\usepackage{subcaption}

\usepackage{multirow}
\usepackage{rotating}
\usepackage{booktabs}
\usepackage{algpseudocode}
\usepackage{algorithm}
\usepackage{etoolbox}
\usepackage{enumitem}

\newtoggle{arxiv}
\toggletrue{arxiv} 

\title{Neuro-Symbolic Generative Art: A Preliminary Study}

\author{Gunjan Aggarwal$^{1}$ and Devi Parikh$^{2}$ \\
guaggarw@adobe.com \quad parikh@gatech.edu\\
$^{1}$Adobe\\
$^{2}$Georgia Tech and Facebook AI Research\\
}

\begin{document} 
\maketitle

\begin{abstract}
\begin{quote}
There are two classes of generative art approaches: neural, where a deep model is trained to generate samples from a data distribution, and ``symbolic'' or algorithmic, where an artist designs the primary parameters and an autonomous system generates samples within these constraints. In this work, we propose a new hybrid genre: neuro-symbolic generative art. As a preliminary study, we train a generative deep neural network on samples from the symbolic approach. We demonstrate through human studies that subjects find the final artifacts and the creation process using our neuro-symbolic approach to be more creative than the symbolic approach 61\% and 82\% of the time respectively. 
\end{quote}
\end{abstract}

\section{Introduction}

Generative art refers to art generated using code, and typically includes an element of chance. Interactive versions of these systems can be viewed as Casual Creators \cite{compton2015casual}. There are two dominant approaches for generative visual art. The first uses deep neural networks to generate images from a distribution that mimics training data. 
Indeed, generative artists train models on specific photographs they take or collect (e.g., Helena Sarin, Robbie Barrat) or perturb the weights of models to create artistic ``glitches'' in the generated art (e.g., Mario Klingemann). Another source of control is the random noise input to the model. Interpolations of two noise vectors smoothly controls the generation in a local neighborhood. 
In the second approach an artist defines an algorithm to generate art. An autonomous system generates random samples using this algorithm. Early algorithmic artists include Georg Nees and Vera Molnar. Algorithmic art is often abstract, with geometric structures, or repeating or recursive patterns. These ``symbolic'' approaches typically have explicit parameters to control the generated art. 
\begin{figure}[t]
\centering
\includegraphics[width=0.95
\columnwidth]{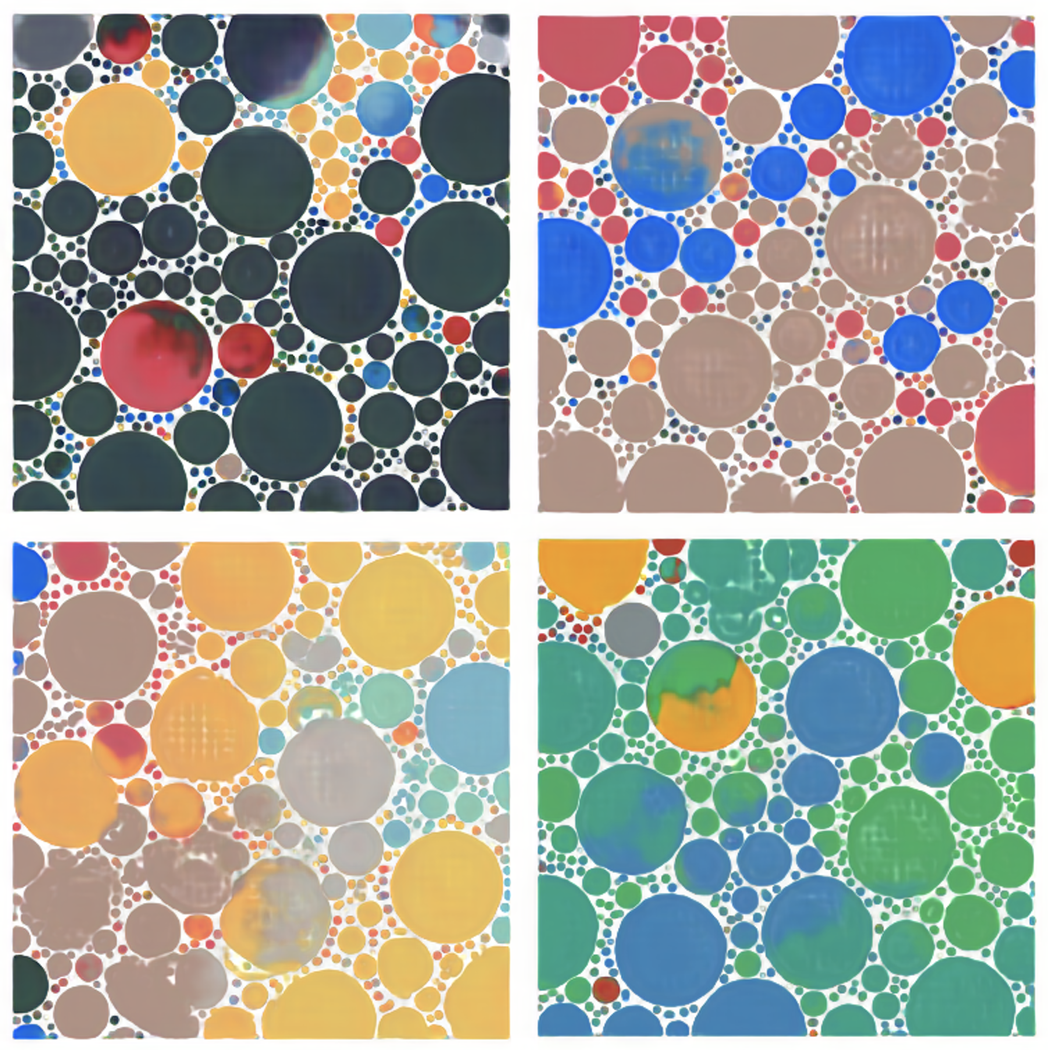}
\caption{Example neuro-symbolic generated art samples.
}
\label{fig:teaser}
\end{figure}

To the best of our knowledge, these two approaches to generative visual art -- neural and symbolic -- have been largely distinct. This work is a preliminary study in exploring their intersection: neuro-symbolic generative art. Specifically, we train a Generative Adversarial Network (GAN) on samples generated using a symbolic approach. We hypothesize that the organic, unpredictable aesthetic associated with neural approaches complements the crisp, designed aesthetic of symbolic approaches. Moreover, compatible with data-hungry deep models, symbolic approaches support generation of large amounts of training samples. Example generated art samples from our approach are shown in Figure~\ref{fig:teaser}. 
Our human studies show that subjects find the artifacts and the interactive creation process using the neuro-symbolic approach to be more creative 61\% and 82\% of the time respectively compared to the symbolic approach.

\section{Related Work}
\noindent \textbf{Neural generative models.} These include Generative Adversarial Networks (GANs) \cite{goodfellow2014generative}, Autoregressive Models \cite{salimans2017pixelcnn++}, Latent Variable Models \cite{kingma2013auto}, etc. Recent progress in GANs enables realistic natural~\cite{brock2018large} and high resolution human face~\cite{karras2019style} image generation. We limit our study to GANs. GANs to generate video game levels~\cite{videoGameGANs} are particularly relevant as neural models trained on procedurally generated content.

\noindent \textbf{Interactive GANs.} Unlike symbolic algorithmic art, GANs do not have  interpretable parameters to control the generated art. The input latent vectors have been shown to frequently contain interpretable variations~\cite{radford2015unsupervised}. \cite{Hrknen2020GANSpaceDI} showed that PCA directions of style latent vectors in StyleGAN ~\cite{karras2019style} contain intuitive interpolation directions such as rotation. \cite{shen2019interpreting} alter facial attributes such as age, pose by editing the latent space of GANs.

\noindent \textbf{Neuro-symbolic AI.} There is currently much debate about the role symbolic reasoning plays in modern AI systems. Neuro-symbolic approaches are considered by some to be at the forefront of the next wave of AI advances. Generative art may serve as a testbed for some of these ideas.

\section{Dataset}
\begin{figure}[t]
\centering
\includegraphics[width=1.0\columnwidth]{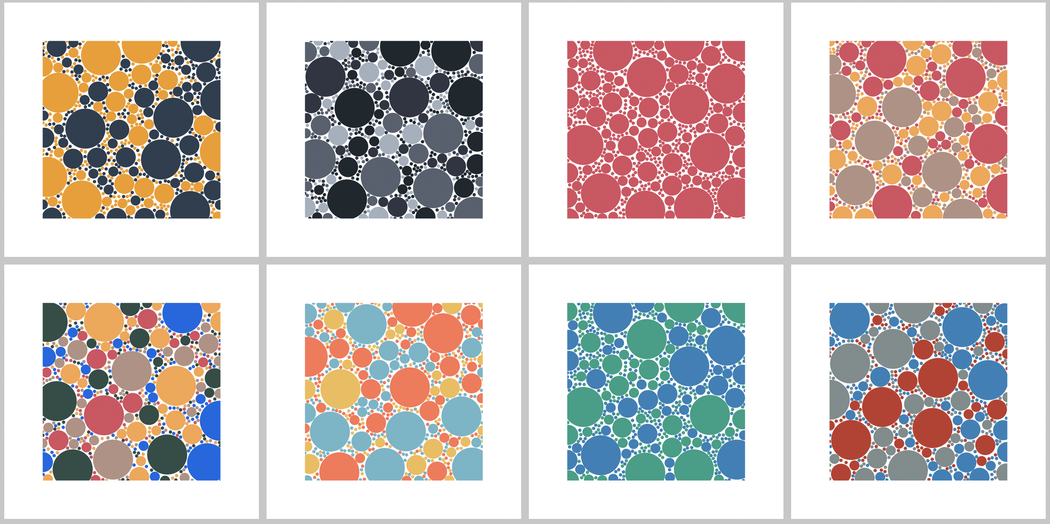}
\caption{Symbolic generated art. Also see Figure \ref{fig:symbolic-more}.
}
\label{fig:symbolic}
\end{figure}
\begin{figure*}[t]
\centering
\includegraphics[width=0.9\textwidth]{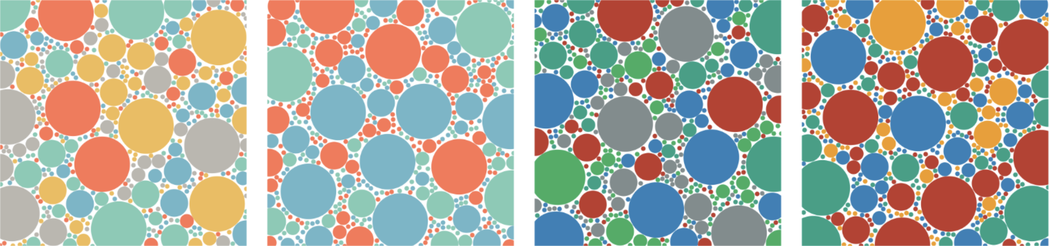}
\bigbreak
\includegraphics[width=0.9\textwidth]{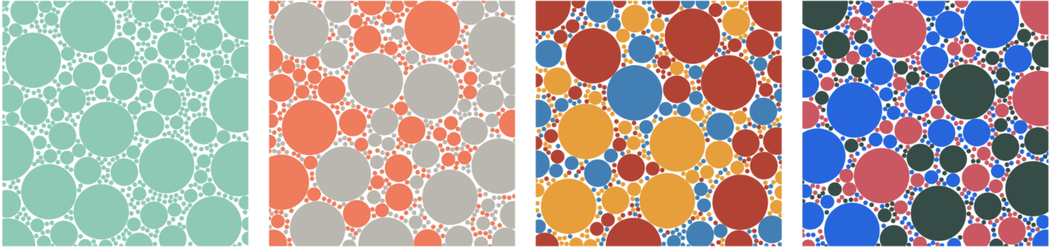}
\bigbreak
\includegraphics[width=0.9\textwidth]{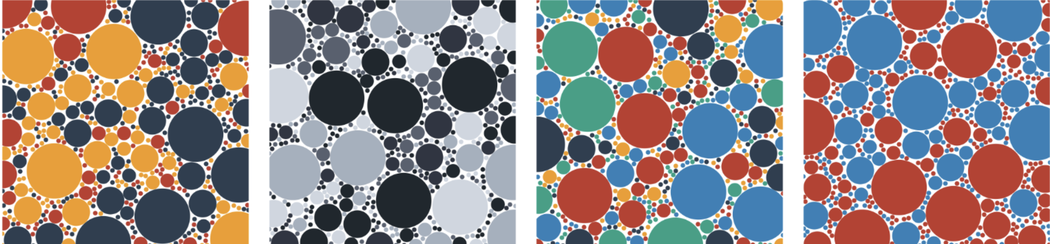}
\bigbreak
\includegraphics[width=0.9\textwidth]{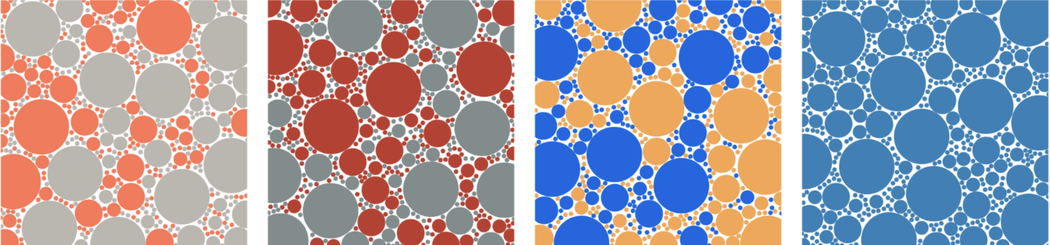}
\bigbreak
\includegraphics[width=0.9\textwidth]{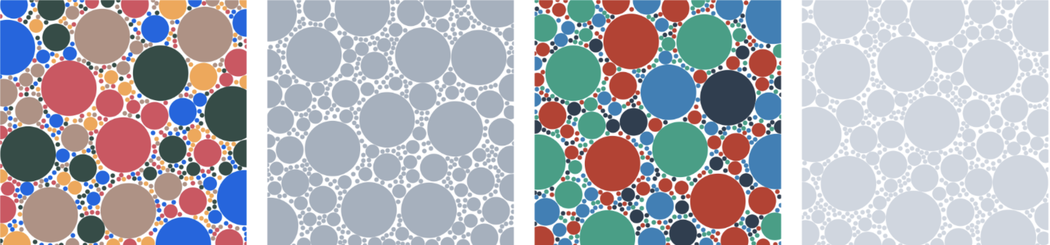}
\bigbreak
\caption{Additional examples of symbolic generated art.
}
\label{fig:symbolic-more}
\end{figure*}

In this preliminary study, we use Circle Packing as our symbolic generator. Non-overlapping circles are placed at random locations in the image. The sizes and number of circles of each size is specified by the artist. Circles are placed in decreasing order of size. The color of each circle is sampled from a  palette. A user can control the art generated via the color palettes (5 options, containing 5 colors each), number of colors sampled from the palette (1 to 5), and the random seed. For each of the 5$\times$5 combinations of the first two, we generate 400 random samples. This results in 10k images in our dataset. See examples in Figures~\ref{fig:symbolic} and~\ref{fig:symbolic-more}.

\section{Approach}

\begin{figure*}[t]
\centering
\includegraphics[width=0.9\textwidth]{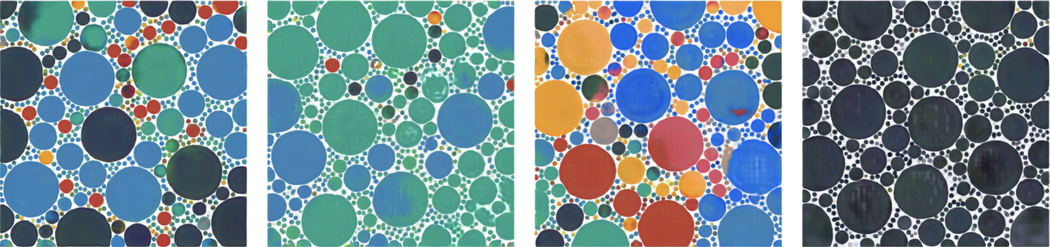}
\bigbreak
\includegraphics[width=0.9\textwidth]{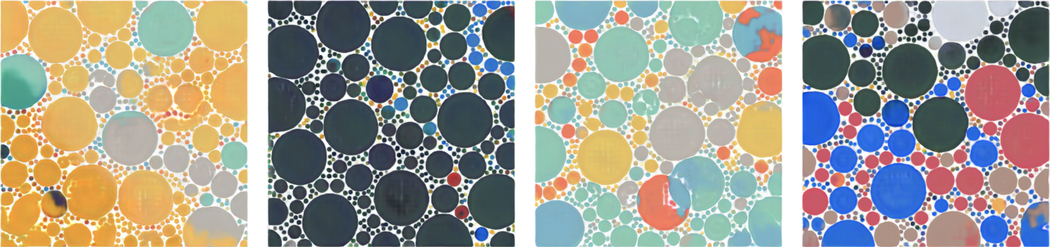}
\bigbreak
\includegraphics[width=0.9\textwidth]{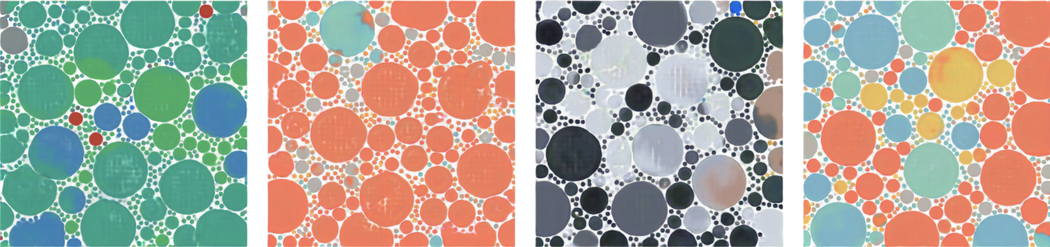}
\bigbreak
\includegraphics[width=0.9\textwidth]{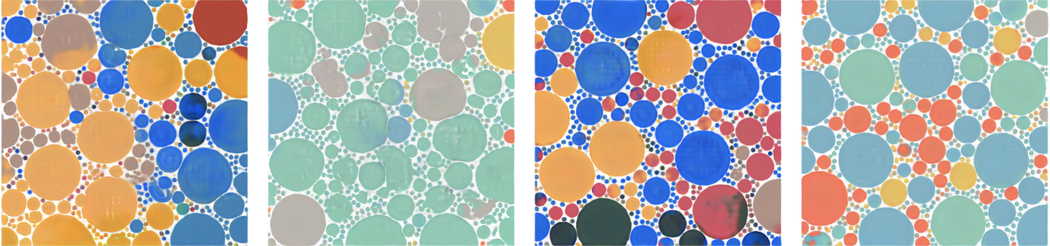}
\bigbreak
\includegraphics[width=0.9\textwidth]{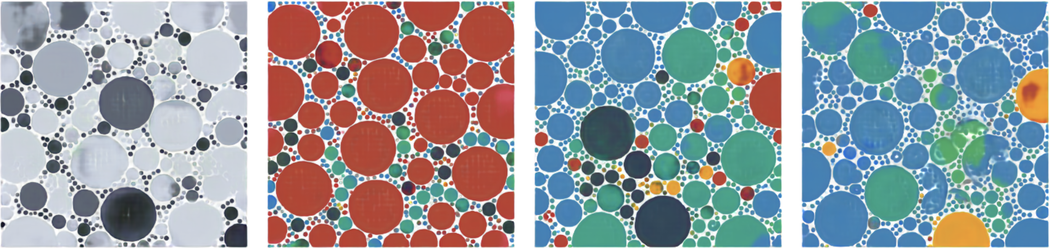}
\bigbreak
\caption{Additional examples of neuro-symbolic generated art.
}
\label{fig:nsg-more}
\end{figure*}

We experimented with different GAN model architectures and found Progressive GAN \cite{karras2017progressive} to work best.
The final generated image is 512$\times$512. The image starts at 4$\times$4, and is  doubled every 37k iterations. The input noise is $512d$. The learning rates for the generator and discriminator were 0.001 for all image resolutions. We use different batch sizes for different image resolutions during training: 128 till 16 $\times$ 16 and then halved after each subsequent increase in resolution. The training is run for 600k iterations with Adam optimizer. We refer to the samples generated from this model as Neuro-Symbolic Generations (NSG). Some examples of NSG can be found in Figure \ref{fig:nsg-more}. Different NSG samples can be generated from the model by feeding it different input noise vectors.

\begin{figure*}[t]
\centering
\includegraphics[width=1.0\textwidth]{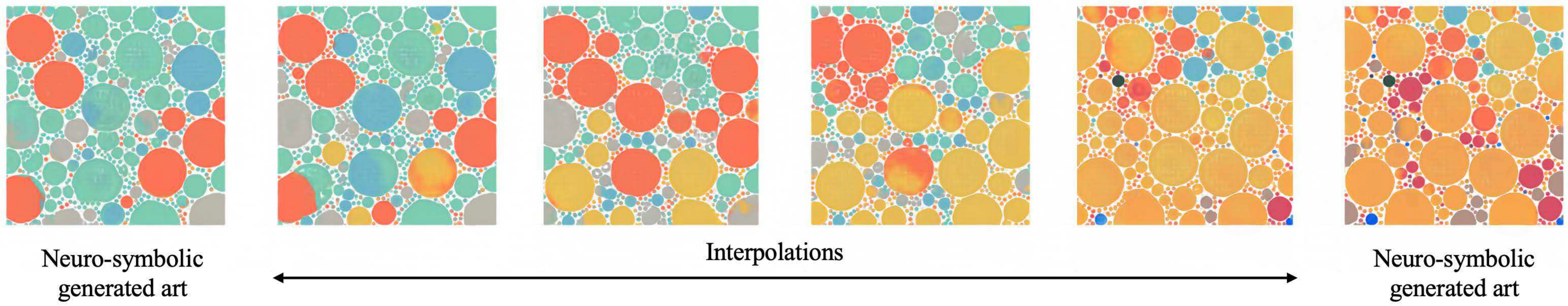}
\caption{Neuro-symbolic interpolations (NSIs) between two neuro-symbolic generations (NSGs). Also see Figure~\ref{fig:nsi-more}.}
\label{fig:interpolations}
\end{figure*}

\begin{figure*}[t]
\centering
\includegraphics[width=1.0\textwidth]{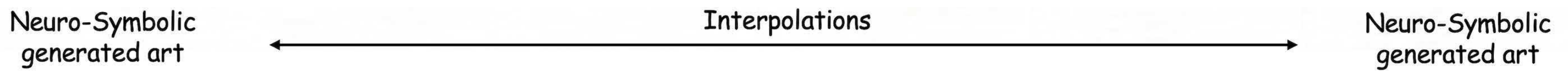}
\includegraphics[width=1.0\textwidth]{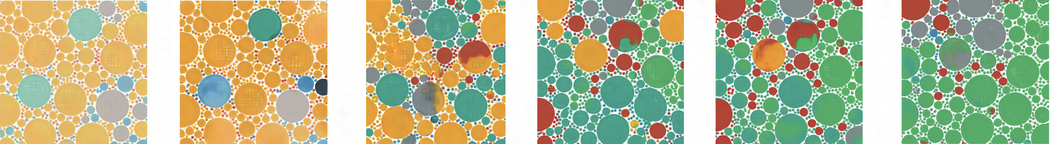}
\bigbreak
\bigbreak
\includegraphics[width=1.0\textwidth]{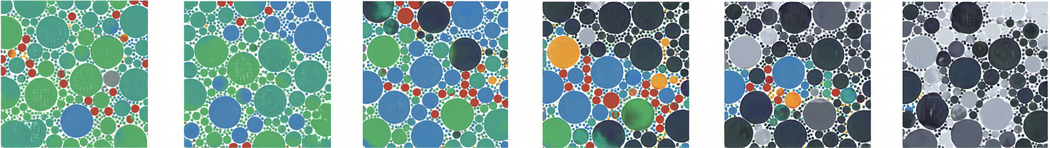}
\bigbreak
\bigbreak
\includegraphics[width=1.0\textwidth]{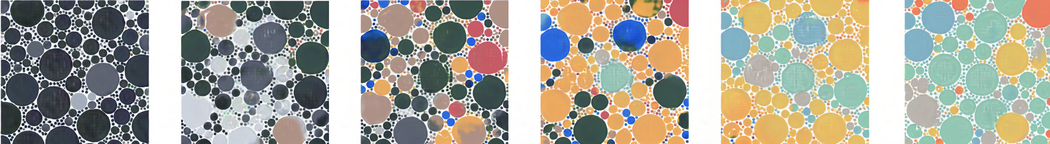}
\bigbreak
\bigbreak
\includegraphics[width=1.0\textwidth]{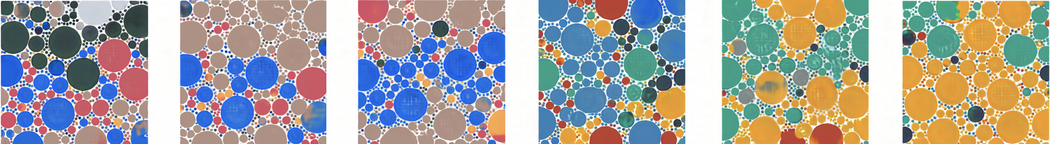}
\bigbreak
\bigbreak
\includegraphics[width=1.0\textwidth]{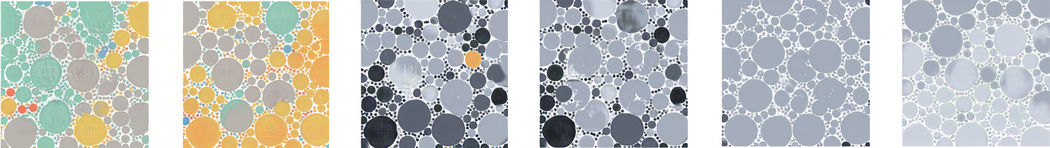}
\bigbreak
\bigbreak
\includegraphics[width=1.0\textwidth]{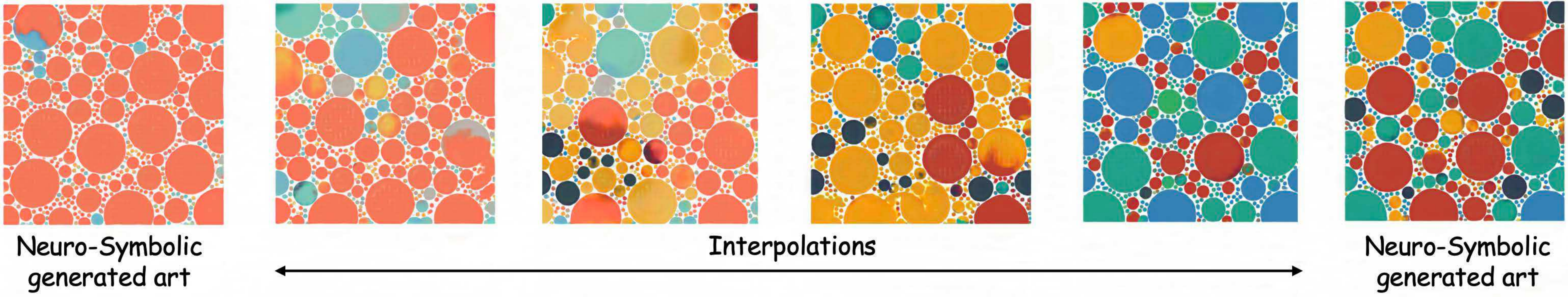}
\bigbreak
\caption{Additional examples of neuro-symbolic interpolations.
}
\label{fig:nsi-more}
\end{figure*}

A user can control the samples via the input noise vector, and interpolations of two noise vectors. For interpolated generation, we sample two noise vectors and then create arbitrary linear interpolations between the two vectors. Neuro-Symbolic Interpolations (NSI) are then generated by feeding the model each of the interpolated latent vectors. Specifically, NSI $x$ is generated from the generative model $G$ as
$$x = G(z);z = z_1 + \alpha*(z_2 - z_1)$$
$$z_1 \sim N(0,1);z_2 \sim N(0,1);\alpha \in (0,1) \text{ set by the user.}$$ Figures~\ref{fig:interpolations} and ~\ref{fig:nsi-more} show example interpolated samples. 
\section{Human Evaluation}

\begin{figure}[t]
\centering
\includegraphics[width=0.9
\columnwidth]{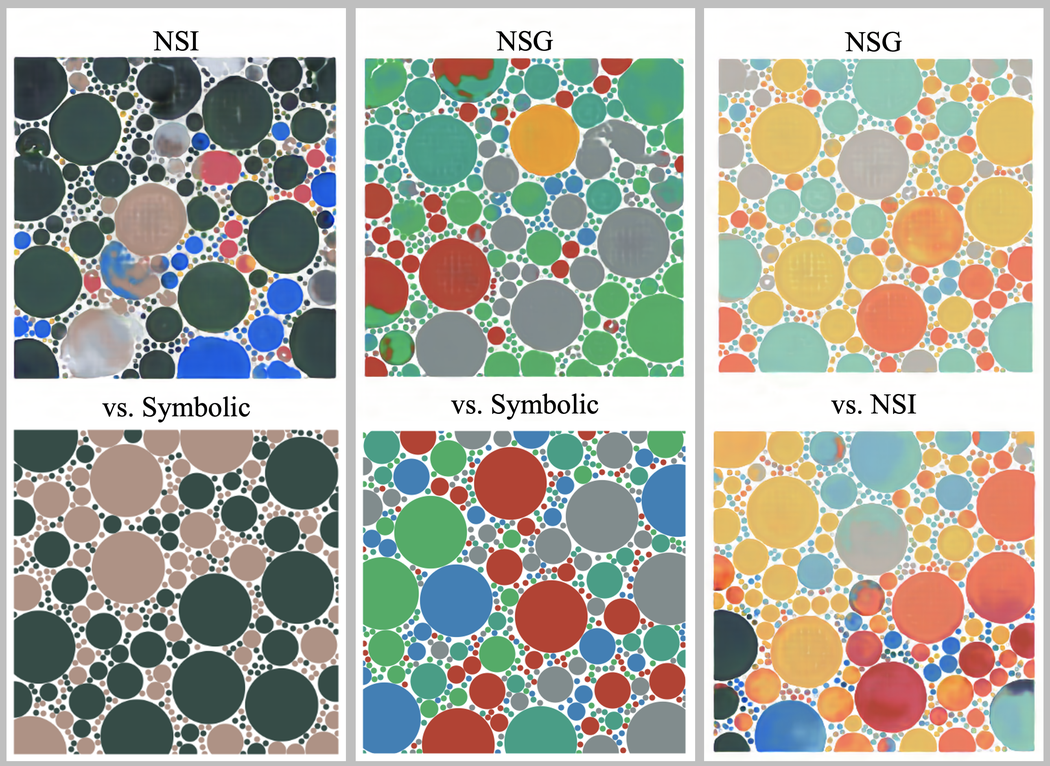}
\caption{Example pairs shown to subjects for evaluation. 
}
\label{fig:artifacts}
\end{figure}

\begin{figure}[t]
\centering
\includegraphics[width=0.95\columnwidth]{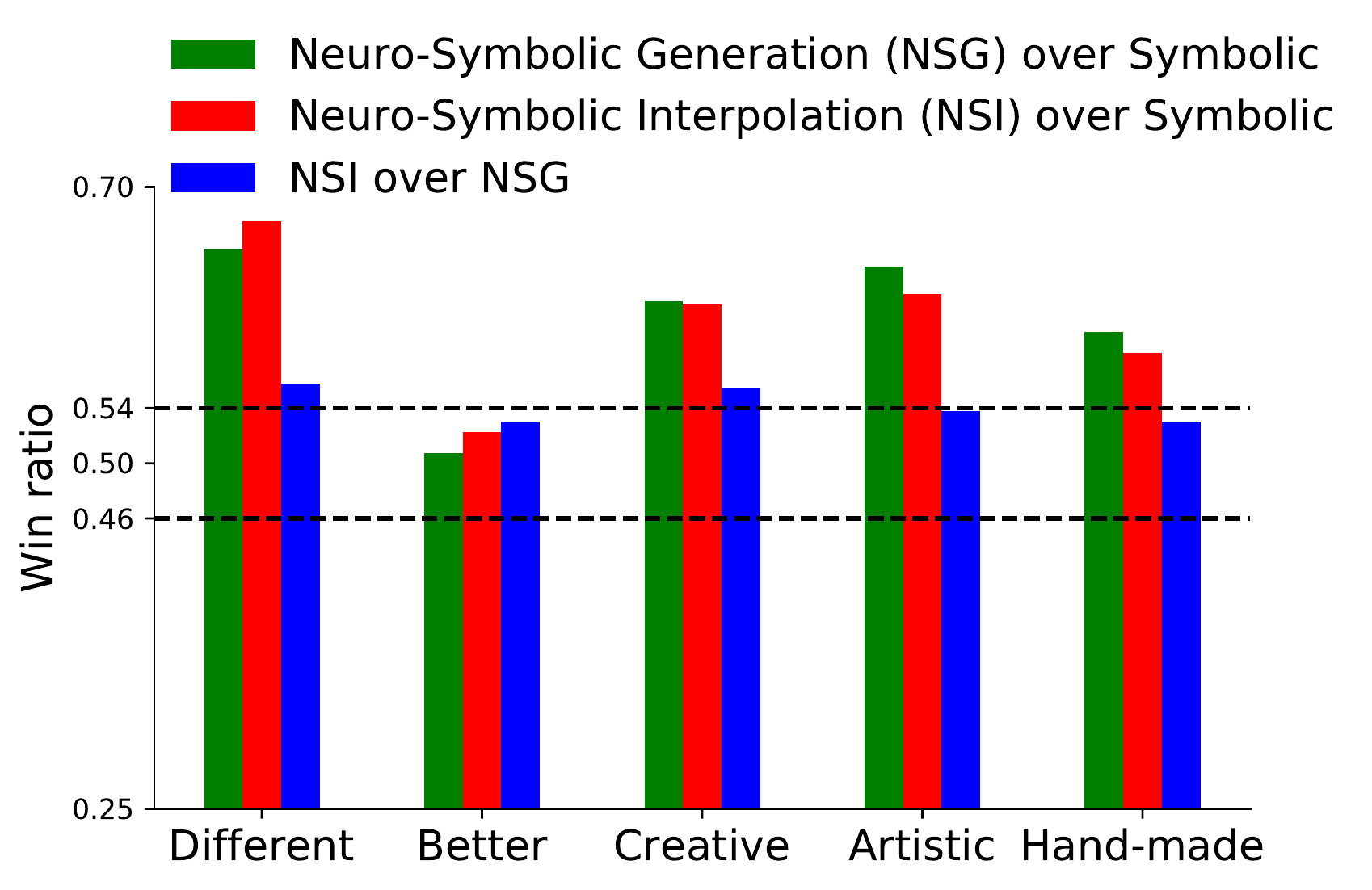}
\caption{Artifact evaluation along 5 axes. Dashed lines denote the band within which a win ratio is not statistically significantly different than 0.5 with 95\% confidence.
}
\label{fig:compare}
\end{figure}
We perform evaluation of both the artifact and the user-driven interactive creation process via human studies on Amazon Mechanical Turk (AMT). Subjects were from the US, with an AMT approval rating of 95\% or higher, and having completed at least 5000 tasks on AMT in the past. They were paid above federal minimum wage.

\subsubsection{Artifact Evaluation}
We compare three artifacts -- Symbolic, Neuro-Symbolic Generation (NSG), and Neuro-Symbolic Interpolated generation (NSI). These replicate the different artifacts a user might create when using the symbolic or neuro-symbolic interactive generative art tools. Human subjects were shown a pair of art pieces, one each from random two of the three types. They were asked which piece
  1) seems more different from art you've seen in your life?
  2) looks better?
  3) is more creative?
  4) is more artistic?
  5) seems more likely to be hand-made?
Subjects were also asked to optionally state why they felt one art was more likely to be hand-made than the other. The study consisted of 60 pairs, equally distributed across the three pairs of approaches. The study was completed by 20 unique subjects, resulting in a total of 1200 pairwise assessments.

Note that if we paired arbitrary pieces from two approaches, more than just the style of the art would likely differ (e.g., the color palette). To control this, the pairs were formed by finding nearest neighbors across approaches using color histograms. This helps minimize unrelated variations, and helps focus the study on the different styles of samples. Example image pairs shown to subjects can be found in Figure~\ref{fig:artifacts}. Specifically, we first generate 10k NSG samples. We then pick a pair of samples and compute an NSI sample associated with the pair using $\alpha$ = 0.5. We generate 10k such NSI samples. Recall that we already have 10k Symbolic samples in our dataset. Now to form a NSG vs. Symbolic pair, we pick either a random NSG or Symbolic sample from our dataset, and find the nearest neighbor from the pool of 10k images of the other category. Same for NSI vs. Symbolic, and NSI vs. NSG. The two images in a pair are randomly shuffled before showing it to subjects.

As quality control, we additionally asked subjects the number of colors in one of the artifacts in a pair. The number of colors in the symbolically generated art is known, giving us a way to identify subjects not doing the task well. Beyond 1 through 5, we gave subjects an added option of ``Shaded colors, so not meaningful to count.''

The proportion of times users preferred one art form over another is shown in Figure~\ref{fig:compare}. 
A one-sample proportion hypothesis test suggests that for our sample size, a ``win ratio'' over 0.54 (or below 0.46) is statistically significant at 95\% confidence. These are shown as a horizontal lines in the figure.  
\textbf{Novelty, unusualness:} We find that the human evaluators rate NSG and NSI as being more ``different'' from art they've seen before than the Symbolic art about 66\% of the time. 
\textbf{Better quality, value:} Subjects like NSG, NSI and Symbolic almost equally.
\textbf{Creativity:} The third and fourth dimensions (``creative'' and ``artistic'') focus directly on the creative aspect of the artifacts. We see that human subjects find NSG and NSI to be more creative than Symbolic art about 61\% and more artistic about 63\% of the times. Note that NS(G/I) and Symbolic rated similarly for quality, but NS(G/I) were rated higher for novelty. We hypothesize that this results in NS(G/I) being rated higher in creativity overall (novelty + value, ~\cite{boden2004creative}).
\textbf{Naturalness, hand-made:} Subjects find NS(G/I) art to be more natural or more likely to be hand-made. Based on the comments shared, while certain subjects preferred Symbolic art as being more hand-made because of \textit{``perfect coloring"}, about 59\% of them chose the NS(G/I) art to be more likely to be hand-made because it
\textit{``Looks like human error with paint dripping on to another color", 
``The other piece of art has solid colors, where as the one I picked has various shades in spots.'',
``mixture of color together'',
``smudge''}. Finally, we see that NSI is preferred over NSG for novelty and creativity.

\subsubsection{Creation Process Evaluation}

\begin{figure}[t]
\centering
\includegraphics[width=0.9
\columnwidth]{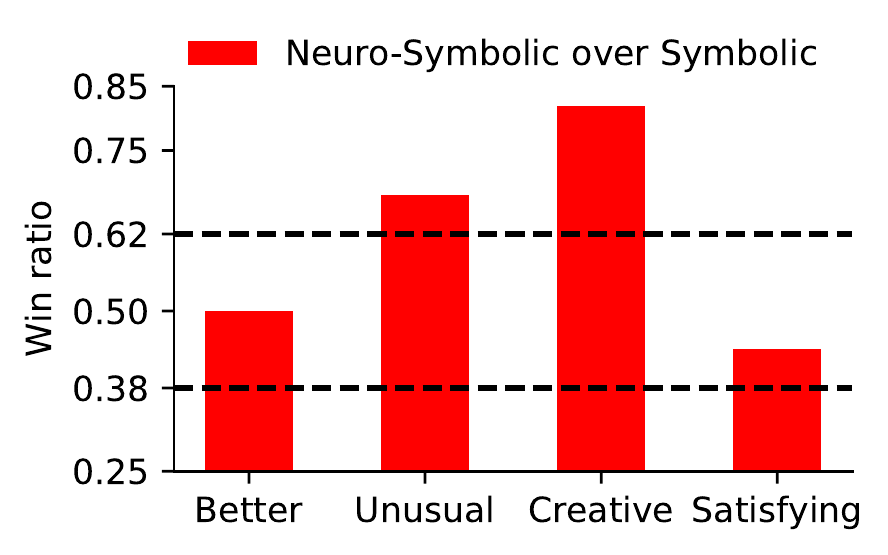}
\caption{Evaluation of the interactive creation process along 4 axes. Win ratios outside the dashed band are statistically significantly different than 0.5 at 95\% confidence. 
}
\label{fig:process-eval}
\end{figure}

Next, we evaluate live interactive generative art tools based on symbolic and our proposed neuro-symbolic approaches. Recall that the symbolic approach has 2 controllable parameters: color palette and the number of colors (maximum 5), as well as an option to generate a new variant of the art with the same parameters by changing the random seed. The neuro-symbolic tool has one controllable parameter $\alpha$ which generates an NSI between 2 NSG pieces, and an option to sample a new pair of NSG art by sampling new noise vectors. Human subjects were given links to both tools (Symbolic: \url{http://genart.cloudcv.org/symbolic}, Neuro-symbolic: \url{http://genart.cloudcv.org}) and for each, were asked to:
  ``Find an art piece that you like a lot and share it with us!'' and 
  describe ``what characteristics of your favorite art made it stand out from others?'' 
Additionally, subjects were asked which tool
1) generates better looking art?
2) generates more surprising / unusual / unpredictable art?
3) generates more creative art?
4) is more satisfying to work with?
50 unique subjects participated. Half were given the symbolic tool first, and the other half the neuro-symbolic tool. Both tools had an option to add up to 5 pieces to their ``favorite'' gallery so users can keep track of pieces they like as they encounter them. Users could delete pieces from the gallery to replace them with others. They were provided an easy way to copy the URL of their favorite piece and submit it.

The proportion of times subjects preferred the neuro-symbolic (NS) tool over symbolic (S) is shown in Figure~\ref{fig:process-eval}. Trends are similar to artifact evaluation. \textbf{Better quality, value:} They like art generated by both tools equally. \textbf{Novelty, unusualness:} Subjects rate NS to be more surprising and unusual than S 68\% of the time. \textbf{Creativity:} Subjects find the NS tool to generate more creative art than S 82\% of the time. \textbf{Satisfying:} Interestingly, while less creative, subjects find S to be more satisfying to work with (albeit, not with statistical significance). An indicative comment from a subject: \textit{``I liked the task. I found that in [NS] the colors felt like they mixed together more. I found that the art in [S] was more clean looking and that made it more satisfying to work with in my opinion."} Using S is perhaps more analogous to ``zen'' (relaxing) activities, while NS may be closer to cognitively taxing creative activities. Exploring this is future work. Other comments about the two tools: \textit{``I had a bit more creative control with [S], while [NS] did generate more interesting combinations, it was just harder to get there predictably.'', ``[NS] provided more creativity, versus taking out colors like in [S].'', ``I liked in [S] the ability to choose the number of colors.  I felt that in [NS] it was a little harder using my mouse to get the form and shape of the circles I wanted.'', ``This was a very interesting experiment, especially [NS]. I kind of felt like I didn't know what to expect when I was trying to make my hybrid art.''}

Example generated art samples beyond those shown in Figures, screen captures of our interactive generative art tools, and ``favorite'' pieces created by subjects along with a description of why they like the pieces can be found here: \url{https://sites.google.com/view/neuro-symbolic-art-gen}.

\section{Conclusion}
We present a preliminary study on neuro-symbolic generative art. It combines what have typically been two distinct approaches to generative visual art: neural and algorithmic/symbolic. We trained GANs on data generated via a symbolic approach. We evaluate the generated art and build live interactive generative art tools using both approaches. Human studies show that subjects find the neuro-symbolic generated art and creation process to be more creative than symbolic counterparts 61\% and 82\% of the time respectively. Overall, we see promising indications that neuro-symbolic generative art may be a viable new genre. 

\noindent \textbf{Future Work.}
We will explore other symbolic art styles, and train a model over multiple styles to potentially discover entirely novel styles. We further plan to interpolate between two symbolic images instead of two neuro-symbolic images. For this, we will explore techniques that map real images to latent representations. A user can then first design the two ends points (symbolically), and then generate an intermediate piece (neurally). Neither symbolic nor neuro-symbolic approaches alone allow for this level of control. Finally, training GANs directly on symbolic representations is an interesting and open research question.
\noindent \textbf{Acknowledgment.}
Abhishek Sinha for helpful discussions.

\bibliographystyle{iccc}
\bibliography{iccc}

\newpage
\vspace{-50mm}

\end{document}